\documentclass{elsart}

\usepackage{amsmath}
\usepackage{amssymb}
\usepackage{graphicx}
\usepackage{rotating}

\makeatletter
\newcommand\figcaption{\def\@captype{figure}\caption}
\newcommand\tabcaption{\def\@captype{table}\caption}
\makeatother

\journal{}

\begin{document}

\begin{frontmatter}

\title{Recognition of expression variant faces using masked log-Gabor features and Principal Component Analysis}

\author{Vytautas Perlibakas}

\ead{vperlib@mmlab.ktu.lt}

\address{
Image Processing and Analysis Laboratory, Computational Technologies Centre, Kaunas University of Technology,
Studentu st. 56-305, LT-51424 Kaunas, Lithuania
}

\begin{abstract}
In this article we propose a method for the recognition of faces with different facial expressions.
For recognition we extract feature vectors by using log-Gabor filters of multiple orientations and scales.
Using sliding window algorithm and variances -based masking these features are extracted at image regions
that are less affected by the changes of facial expressions. Extracted features are passed to the
Principal Component Analysis (PCA) -based recognition method.
The results of face recognition experiments using expression variant faces showed that the proposed method
could achieve higher recognition accuracy than many other methods. For development and testing we used facial
images from the AR and FERET databases.
Using facial photographs of more than one thousand persons from the FERET database the proposed method
achieved 96.6-98.9\% first one recognition rate and 0.2-0.6\% Equal Error Rate (EER).
\end{abstract}

\begin{keyword}
Face recognition \sep Principal Component Analysis \sep Log-Gabor filters \sep Facial expression
\end{keyword}

\end{frontmatter}

\section{Introduction}

Face recognition is a difficult problem, because the accuracy of recognition can be affected by many factors:
different environment and lighting conditions, different input devices and their parameters, changes of the
face itself (due to the change of expression, make-up, face rotation or aging).
In this article we investigate the problem of recognizing faces with different facial expressions.

Traditional greyscale pattern -based recognition methods (like Principal Component Analysis (PCA) \cite{Turk1991})
do not solve the problem of recognizing faces with different expressions.
This problem becomes obvious if we perform training and store in the database faces with one expression
(for example, neutral) and then need to recognize faces with another expression (for example, smiling).
The accuracy of recognition in such case is usually lower than the accuracy of recognizing faces with
the same expressions.
In order to improve the recognition accuracy of expression variant faces we can detect some facial features
(for example, around lips, eyes, nose) and then compare these corresponding features of different faces.
Such feature detection is used in the Elastic Bunch Graph Matching (EBGM) and Gabor wavelets -based face recognition method that
was developed by Wiskott and his colleagues \cite{Wiskott1997}.
Penev and Atick \cite{Penev1996} developed Local Feature Analysis (LFA) -based face recognition algorithm that
detects features (not necessarily related with anthropometrical features like eyes or lips) in any part of the face
and uses them for recognition.
Expression variant faces also can be compared by using optical flow -based algorithm \cite{Moghaddam2001},
and the accuracy of recognition can be improved by appropriately weighting
image pixels, regions or features that are more or less affected by the changed expression \cite{Martinez2003}.

In this article we investigate how the accuracy of face recognition is affected by the change of facial expression
and propose a masked log-Gabor features and Principal Component Analysis -based method that could increase the accuracy
of recognizing faces with different expressions. Extensive recognition experiments using a large database of facial images
and comparison with the results of other researchers showed that our method achieves very high recognition accuracy.

\section{Face recognition using masked log-Gabor features and PCA}

\subsection{Feature extraction using log-Gabor filters and sliding window algorithm}

For face recognition we used greyscale facial images and performed feature extraction by using 
the log-Gabor filters that were proposed in \cite{Field1987} for coding
of natural images. The experiments showed, that these filters are more suitable for image coding than the traditional Gabor filters.
The log-Gabor filter in the frequency domain and polar coordinates can be calculated using the following equation \cite{Kovesi1996}:
\begin{equation}
G(f,\theta ) = \exp \left( { - \frac{{\ln ^2 \left( {{f \mathord{\left/
 {\vphantom {f {f_0 }}} \right.
 \kern-\nulldelimiterspace} {f_0 }}} \right)}}
{{2 \cdot \ln ^2 \left( {{k \mathord{\left/
 {\vphantom {k {f_0 }}} \right.
 \kern-\nulldelimiterspace} {f_0 }}} \right)}}} \right) \cdot \exp \left( { - \frac{{\left( {\theta  - \theta _o } \right)}}
{{2 \cdot \sigma _\theta ^2 }}} \right)
\end{equation}
here
$f_0$ is the centre frequency of the filter, $k$ determines the bandwidth of the filter, 
$\theta _o$ is the orientation angle of the filter, and
$\sigma _\theta   = {{\vartriangle \theta } \mathord{\left/
 {\vphantom {{\vartriangle \theta } {s_\theta  }}} \right.
 \kern-\nulldelimiterspace} {s_\theta  }}$
where $s_\theta$ - scaling factor, $\vartriangle \theta$ - orientation spacing between filters.
For face recognition we generated multiple log-Gabor filters $G_{n_{o}, n_{s}}$ of different orientations $n_{o}$ and scales $n_{s}$ using the following
relationships:
$f_0  = {1 \mathord{\left/
 {\vphantom {1 \lambda }} \right.
 \kern-\nulldelimiterspace} \lambda }$,
$\lambda  = \lambda _0  \cdot s_\lambda ^{(n_s  - 1)} $,
${k \mathord{\left/
 {\vphantom {k {f_0 }}} \right.
 \kern-\nulldelimiterspace} {f_0 }} = \sigma _f $,
$\sigma _f (\beta ) = \exp ( - 0.25\beta \sqrt {2 \cdot \ln (2)} )$ \cite{Boukerroui2004},
$n_s  = 1,...,N_s$;
$\theta _o  = {{\pi (n_o  - 1)} \mathord{\left/
 {\vphantom {{\pi (n_o  - 1)} {N_o }}} \right.
 \kern-\nulldelimiterspace} {N_o }}$,
$\vartriangle \theta  = {\pi  \mathord{\left/
 {\vphantom {\pi  {N_o }}} \right.
 \kern-\nulldelimiterspace} {N_o }}$,
$n_o = 1,...,N_o$.
Here 
$\lambda _0$ is the wavelength of the smallest scale filter,
$s_\lambda$ is the scaling factor between successive filter scales,
$\beta$ is the bandwidth of the filter in octaves,
$N_s$ is the number of scales,
and $N_o$ is the number of orientations.
Using a chosen two-dimensional log-Gabor filter $G_{n_{o}, n_{s}}$ in Fourier space
we perform filtering (convolution), magnitude calculation and
masking:
\begin{equation}
V_{n_{o}, n_{s}}=abs(IFFT2(G_{n_{o}, n_{s}}.*FFT2(I))).*mask,
\end{equation}
here
"$.*$" - array multiplication,
$I$ - normalized (derotated, cropped, resized, masked) face image,
$G_{n_{o}, n_{s}}$ - log-Gabor filter of a chosen orientation and scale in Fourier space (the size of the filter array is the same as the size of the two-dimensional image $I$),
$FFT2$ - two-dimensional Fast Fourier Transform, $IFFT2$ - inverse $FFT2$,
$mask$ - binary mask for masking log-Gabor magnitude image (the same as is used for masking greyscale face image $I$ in order
to leave only the internal part of the face), $V_{n_{o}, n_{s}}$ - masked log-Gabor magnitude image.
It must be noted that for the selected image size we can pre-calculate the log-Gabor filters
(of different sizes and orientations) only once and store.

\begin{figure}[htbp]
\centering \includegraphics[width=120mm]{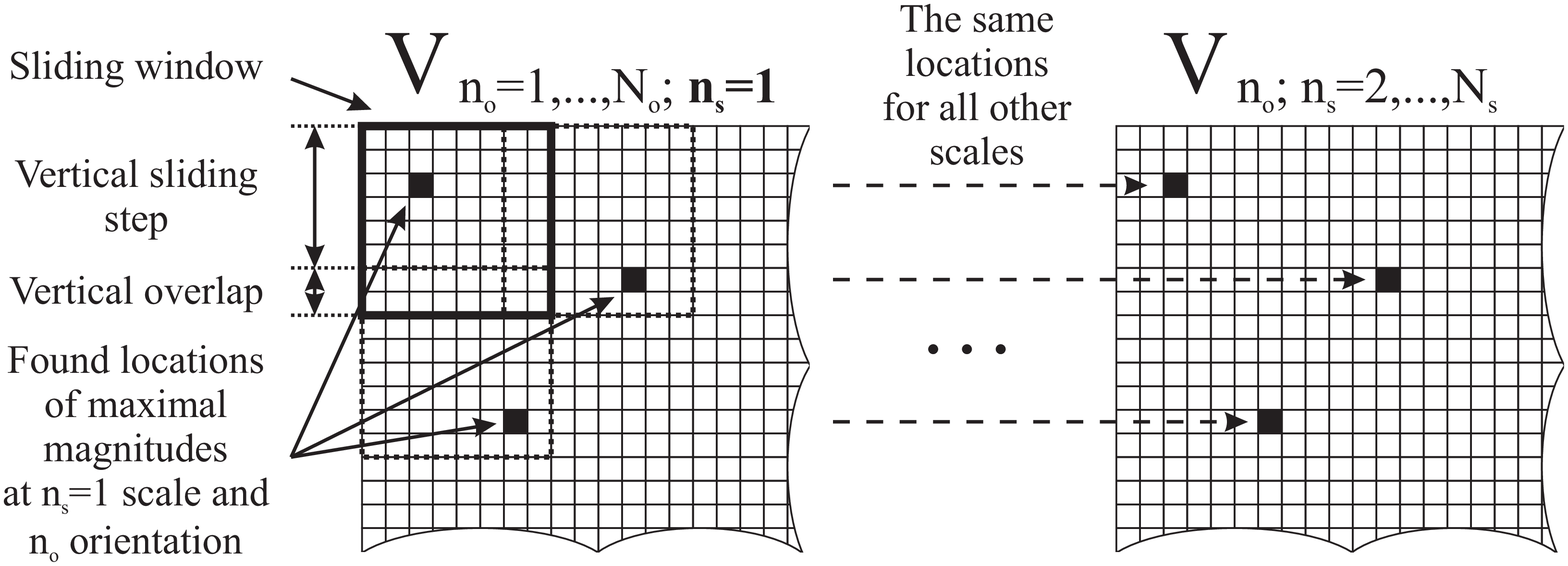}
\caption{Selection of Log-Gabor magnitude features using sliding window algorithm.}
\label{fig_sliding_window}
\end{figure}

\begin{figure}[htbp]
\begin{minipage}[b]{0.5\linewidth}
\centering \includegraphics[width=70mm]{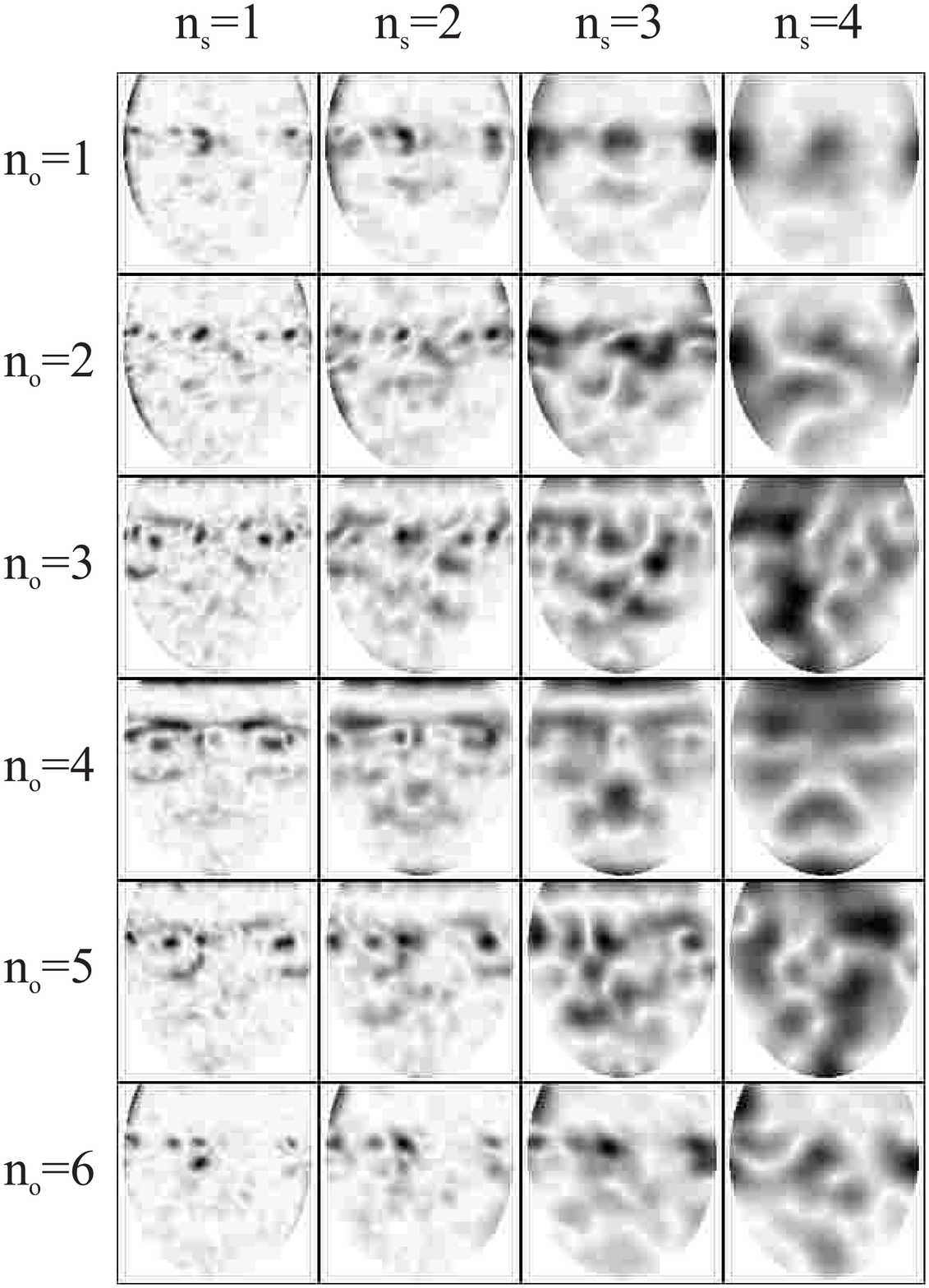}
\end{minipage}%
\hspace{0.5cm}
\begin{minipage}[b]{0.5\linewidth}
\centering \includegraphics[width=70mm]{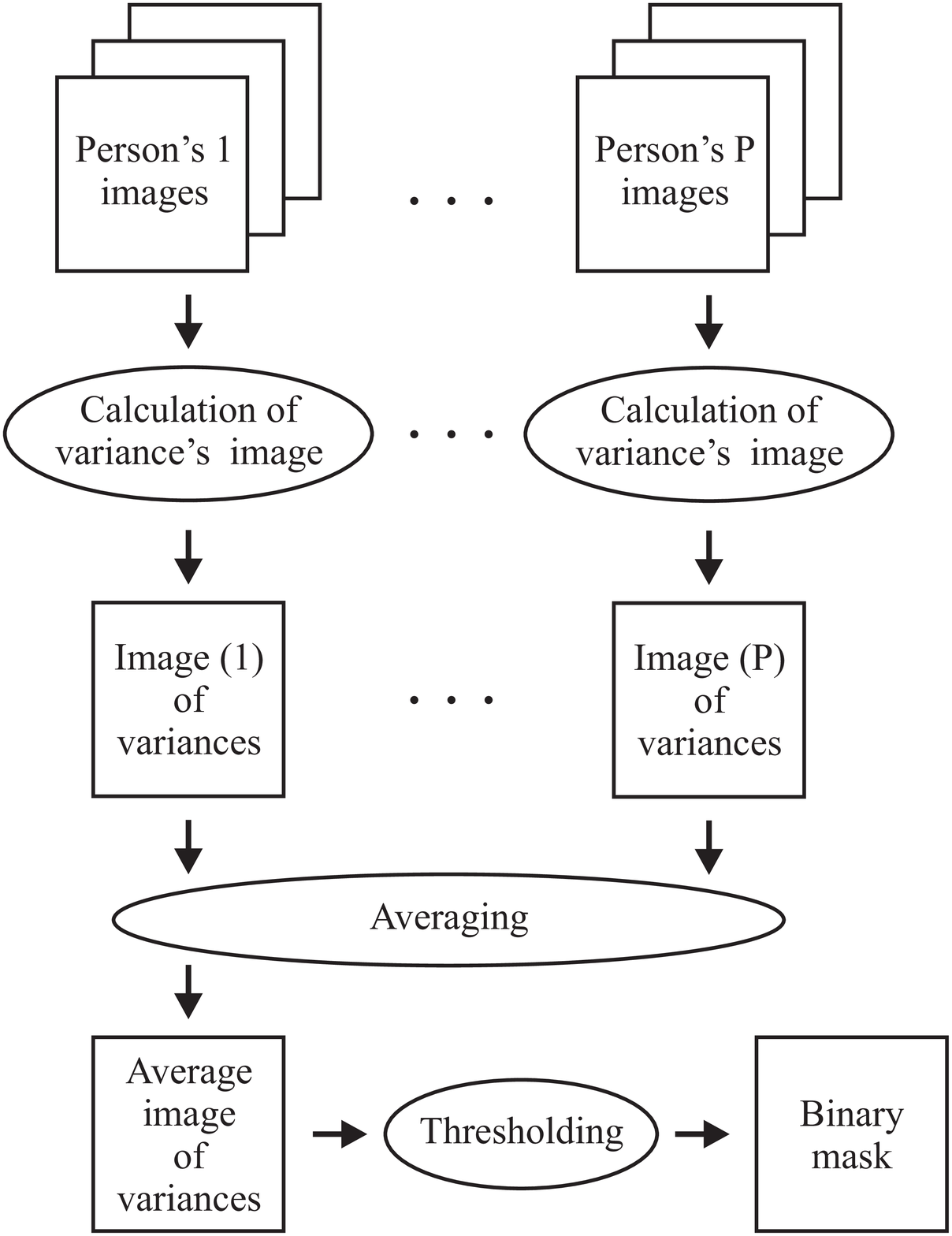}
\end{minipage}

\begin{minipage}[t]{0.5\linewidth}
\caption{Log-gabor magnitude images of $N_{o}=6$ orientations and $N_{s}=4$ scales (dark regions mean high magnitudes).}
\label{fig_magnitudes}
\end{minipage}%
\hspace{0.5cm}
\begin{minipage}[t]{0.5\linewidth}
\caption{Calculation of variances and masking images.}
\label{fig_variances}
\end{minipage}

\end{figure}

After image filtering with multiple log-Gabor filters ($N_{s}$ scales and $N_{o}$ orientations) we get a very
large number of log-Gabor features (magnitude values in all $N_{s} \cdot N_{o}$ magnitude images as it is shown in Fig. ~\ref{fig_magnitudes}).
The size of each magnitude image $V$ is the same as the size of facial image $I$.
In order to reduce the number of features we use sliding window algorithm that is illustrated in Fig. ~\ref{fig_sliding_window}.
Rectangular window of a chosen size is slided over the magnitude image $V_{n_{o}, 1}$ (scale $n_{s}=1$) using
a chosen sliding steps.
In each window we find one maximal magnitude value and remember
the location (coordinates in image $V_{n_{o}, 1}$) of this value.
Features at all other scales $n_s  = 2,...,N_s$ of the same orientation $n_{o}$ are extracted at the same locations
without using sliding window as it is shown in Fig. ~\ref{fig_sliding_window}.
The same feature's finding procedure is repeated for all $N_{o}$ orientations.
When we perform face recognition, the log-Gabor features (found using sliding window)
for each image from the database of faces are calculated only once and stored.
If the magnitude image is masked, we perform search only in an unmasked image part.
Then all extracted log-Gabor features (magnitude values)
are stored in a one-dimensional vector $X$ and passed to the Principal Component Analysis -based recognition method.

\begin{figure}[htbp]
\centering \includegraphics[width=140mm]{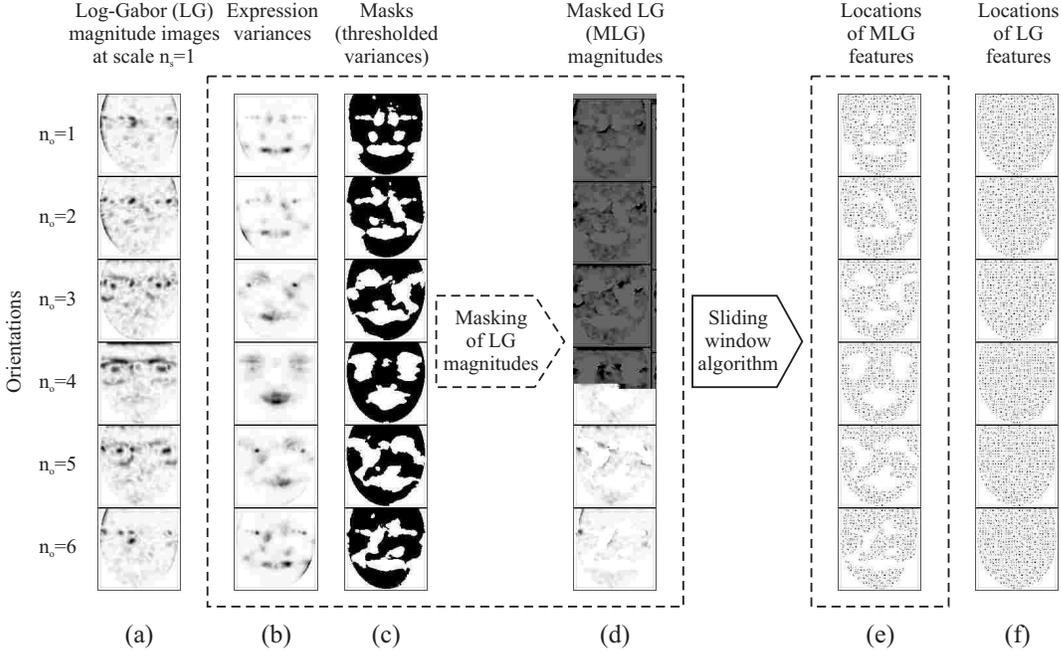}
\caption{
Features masking procedure: (a) Log-Gabor (LG) magnitude images at scale $n_s=1$;
(b) Pre-calculated variance images (dark regions mean high variances);
(c) Feature masks (expression masks) that were calculated by thresholding variance images at average levels;
(d) Masked log-Gabor (MLG) magnitude images;
(e) MLG feature locations that are found using sliding window algorithm (locations are denoted by black points);
(f) LG feature locations that are found if we omit feature's masking steps (b)-(d).
}
\label{fig_masking_procedure}
\end{figure}

In order to increase the accuracy of recognizing expression variant faces we used masking of log-Gabor magnitude images.
Using masks we eliminated the regions of magnitudes that were mostly affected by changed facial expressions.
These changes were evaluated by calculating magnitude variances between images with different expressions
of the same person. Variance images of different persons were averaged and thresholded (by average). After thresholding obtained
binary image was used for masking.
For calculation of variance images and masks we use not greyscale images, but the log-Gabor magnitude images
at first scale. We calculate masks for each filter orientation.
Masks were created for scales $n_{s}=1$ and then used for all scales $n_s  = 1,...,N_s$ of the same orientation.
The number of masks is the same as the number of orientations.
Mask calculation procedure is illustrated in Fig. \ref{fig_variances}.
Masking of log-Gabor magnitude images is illustrated in Fig. \ref{fig_masking_procedure}.
After masking we use sliding window algorithm and find the log-Gabor features in an unmasked parts of these masked images.
Found features are used for PCA -based recognition.

\subsection{Face recognition using PCA of log-Gabor features}

In this section we will present the main formulas of the Principal Component Analysis (PCA) -based recognition method.
More information about PCA could be found in \cite{Kirby1990}.

In order to perform PCA, we calculate the covariance matrix \hbox{ $C=\frac{1}{r}\sum\limits_{j = 1}^r d_j d_j^T$ },
here centred data vector $d_j=X_j-m$, mean vector $m=\frac{1}{r}\sum\limits_{j = 1}^r {X_j}$,
data vector $X=(x_1,x_2,...,x_N)^T$, the number of data values is $N$, the number of data vectors is $r$.
Then we find the eigenvectors $u_k$ and eigenvalues $\lambda_k$ of this covariance matrix: $Cu_k=\lambda_k u_k$.
Eigenvectors that correspond to the largest eigenvalues are used to construct the transformation matrix $T$.
This matrix is used to transform any data vector $X$ to the PCA space: 
$
Y = \Lambda ^{ - {1 \mathord{\left/
 {\vphantom {1 2}} \right.
 \kern-\nulldelimiterspace} 2}} T(X - m),
$
here $\Lambda$ is a diagonal matrix (of the corresponding eigenvalues)
that performs data "whitening" \cite{Bishop1995}.

Face recognition is performed by comparing feature vectors $Y$ of different facial images. For comparison
we used the cosine -based distance measure $d=-cos(Y_1,Y_2)$, here $Y_1$ and $Y_2$ are PCA
feature vectors of the compared facial images $X_1$ and $X_2$. Unknown faces are rejected
by using some manually chosen distance's threshold. When we perform traditional PCA -based face recognition,
data vector $X$ contains greyscale values of the image pixels. When we perform log-Gabor PCA -based recognition, this
vector contains the selected log-Gabor magnitude values (using expression masking and sliding window -based algorithm).

\section{Face recognition experiments and results}

\subsection{Recognition performance measures}

\begin{figure}[htb]
\centering \includegraphics[width=140mm]{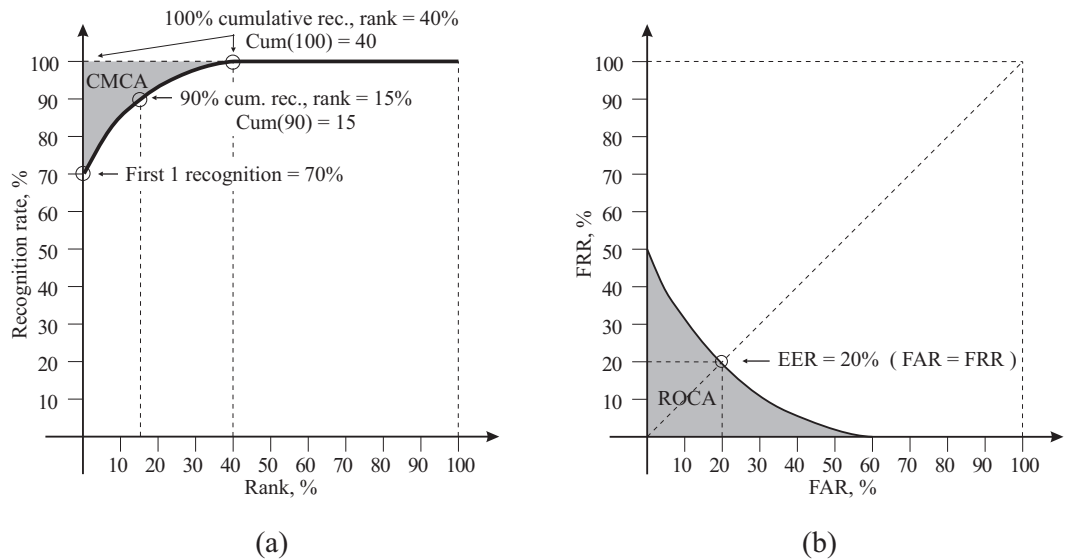}
\caption{
Recognition performance characteristics: (a) Cumulative Match Characteristic (CMC); (b) Receiver Operating Characteristic (ROC).}
\label{fig_cmc_roc}
\end{figure}

For comparison of face recognition methods we used
Cumulative Match Characteristic (CMC) and Receiver Operating Characteristic (ROC) - based measures
that we also used in \cite{Perlibakas2004a}.
We used the following measures:
1) First one recognition rate (First 1, $[0,100\%]$). For example, First 1 = 70\% means that 
in 70 \% cases the persons (that we are looking for in the database) will be identified correctly
by extracting from the database one most similar photography. Larger First 1 values mean better result.
2) Rank (number) of images (Cum, $(0,100\%]$) that we must extract from the database in order
to achieve some cumulative recognition rate. For example, Cum(90\%)=15\% means that we need to extract
15\% of images from the database in order to achieve not smaller than 90\% cumulative recognition rate.
That is, the faces that we look for appear between extracted 15\% of images (not necessarily first one) in 90\% cases.
Smaller ranks (at the same recognition rate) mean better result.
3) The area (CMCA, $[0,100^4]$) above Cumulative Match Characteristic (CMC) that measures overall identification accuracy.
Smaller CMCA means better result.
4) The area (ROCA, $[0,100^4]$) below Receiver Operating Characteristic (ROC) that measures overall verification accuracy.
Smaller ROCA means better result.
5) Equal Error Rate (EER, $[0,100\%]$). For example, EER=20\% means that FAR=FRR=EER=20\%, in 20\% cases 
a non authorized person was accepted as authorized (False Acceptance Rate), and also in 20\% cases 
an access was not granted (rejected) for an authorized person (False Rejection Rate). Smaller EER means better result.
Graphical representation of the used characteristics and measures is presented in Fig. ~\ref{fig_cmc_roc}.

\subsection{Face normalization and feature extraction}

\begin{figure}[htbp]
\centering \includegraphics[width=120mm]{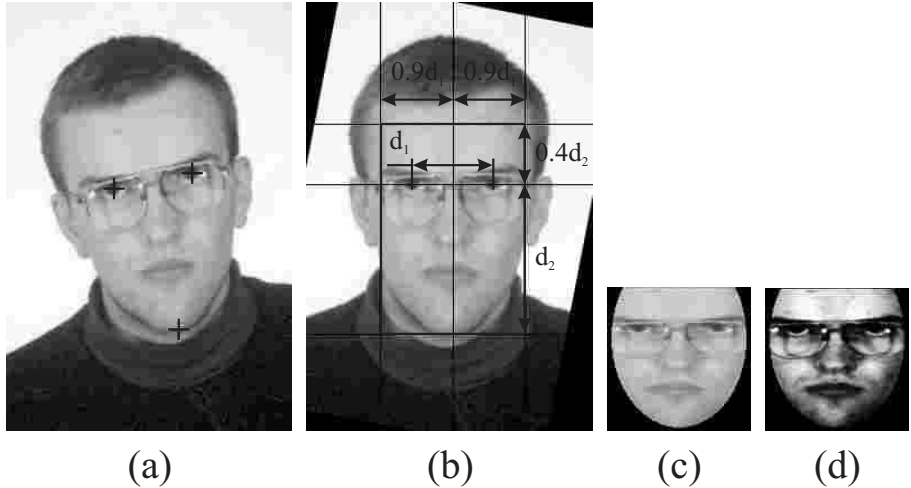}
\caption{Image normalization: (a) Initial image with selected eyes and chin; (b) Denoised, derotated image, and cropping schema; (c) Cropped, resized and masked image; (d) Normalized image after histogram equalization.}
\label{fig_image_normalization}
\end{figure}

For recognition we used greyscale facial images with manually selected centres of eyes and the tip of chin.
Initial images were denoised using Gaussian filter with $\sigma=0.5$ and window size 5x5 pixels.
Then using manually selected points we performed geometrical normalization.
Faces were derotated (in order to make the line connecting eye centres horizontal), cropped, resized (to the size of 128x128 pixels),
masked. For rotation and resizing we used bicubic interpolation.
For masking we used an ellipse with 
central point (64.5,45.5), horizontal axis of 120 pixels, and vertical axis of 160 pixels.
Then for an unmasked part of the image we performed histogram equalization (256 levels).
Image normalization procedure is presented in Fig. ~\ref{fig_image_normalization}.
For illustration we used an image from our personal archive. The same face normalization procedure was used
in all our recognition experiments.

After image masking using elliptical mask are left 12646 unmasked pixels (grey value features) of 16384 (128x128).
These grey value features are used for traditional PCA -based recognition method.
For log-Gabor PCA (LG PCA) we leave (using sliding window) 19704 log-Gabor magnitude features.
Sliding window size is 4x4 pixels, sliding step is 4 pixels.
For masked log-Gabor PCA (MLG PCA) we leave 15120 log-Gabor magnitude features (total at all orientations and scales).
Log-Gabor filters were generated using the following parameters:
$\lambda _0 = 5$,
$s_\lambda = 1.6$,
$\beta = 1$,
$\sigma _f (\beta) = 0.745$,
$N_s  = 4$,
$s_\theta = 1.5$,
$N_o = 6$.
Log-Gabor magnitude images were masked using the same elliptical mask that was used for image normalization.
For calculation of expression masks we used facial images of 117 persons from the AR \cite{Martinez1998} database (first session).
We used facial images with three expressions (neutral, smiling, angry) per person.
Face recognition experiments were performed using the following 3 methods:
1) PCA with grey value images (notation "PCA");
2) PCA with log-Gabor features and without expression masks (notation "LG PCA");
3) PCA with masked log-Gabor features (notation "MLG PCA").
All these methods used the same normalized image patterns and the same implementation of PCA with
cosine -based distance measure between "whitened" feature vectors.
For face recognition experiments we used the AR \cite{Martinez1998} and FERET \cite{Phillips1998} databases.

\subsection{Recognition experiments using AR database}

The AR face database \cite{Martinez1998} was created by A. Martinez and R. Benavente at Computer Vision Center, Purdue University
in 1998.  This database contains facial photographs of 126 persons with strictly controlled facial expressions and lighting.
The size of images is 768x576 pixels.
Images of each person were captured in two sessions that were separated by two weeks time.
From this database we selected images of 117 persons (64 men, 53 women) with the following three controlled
expressions: neutral expression (NE), smile (SM), anger (AN).
Images with expression numbers 1, 2, 3 of the first session we denote by NE1, SM1, AN1, and images
with expression numbers 14, 15, 16 of the second session we denote by NE2, SM2, AN2.
Facial images of the first session were used for training and registration (known faces) and 
for creation of variances -based expression masks.
Facial images of the second session were used as an unknown faces that we wish to recognize.
All images we converted to greyscale format. For recognition we used 100 PCA features.

At first we performed recognition experiments using NE1, SM1, AN1 images for training and registration, and
using NE2, SM2, AN2 as an unknown faces.
The results of experiments showed (Table \ref{tbl_rec_ar1}),
that the best recognition accuracy is achieved if we can ensure that the expressions
of training, registered, and unknown faces will be the same. Also we can see that the worst recognition
accuracy is achieved in the following cases: 1. If we use smiling (SM1) faces for training and registration,
and then wish to recognize faces with neutral (NE2) or angry (AN2) expressions; 2. If we use faces with neutral (NE1)
or angry (AN1) expressions for training and registration, and then wish to recognize faces with smiling expressions (SM2).
In such hard cases LG PCA and MLG PCA methods allow to achieve 10-38\% higher first one recognition accuracy than
traditional PCA. In these cases MLG PCA achieves slightly higher recognition accuracy than LG PCA.

\begin{sidewaystable}

\caption{Recognition using AR database and the same expressions for training and registration (known faces).}
\label{tbl_rec_ar1}

\begin{tabular}{ccccccccccc}
\hline
Known	& Method	&\multicolumn{9}{c}{Unknown faces} \\
\cline{3-11}
faces	&			&\multicolumn{3}{c}{NE2} & \multicolumn{3}{c}{SM2} & \multicolumn{3}{c}{AN2} \\
\cline{3-11}
		&			& First 1 & CMCA   & ROCA	& $\mid$ First 1 & CMCA	  & ROCA   & $\mid$ First 1	& CMCA	 & ROCA \\
\hline
NE1		& PCA		& 94.87   & 161.44 & 63.01	& 68.38			 & 264.08 & 169.25 & 80.34			& 228.29 & 123.36 \\
		& LG PCA	& 99.15	  & 91.68  & 2.07	& 87.18			 & 142.82 & 51.37  & 96.58			& 94.97	 & 5.76 \\
		& MLG PCA	& 98.29	  & 86.20  & 0.94	& 93.16			 & 113.96 & 23.73  & 97.44			& 89.49	 & 1.74 \\
\hline
SM1		& PCA		& 63.25	  & 321.79 & 231.94 & 96.58			 & 89.85  &	2.47   & 45.30			& 618.01 & 528.01 \\
		& LG PCA	& 83.76	  & 184.45 & 96.67  & 99.15			 & 85.84  &	0.96   & 70.94			& 211.85 & 114.96 \\
		& MLG PCA	& 92.31	  & 126.74 & 31.39  & 99.15			 & 86.57  &	0.79   & 83.76			& 139.89 & 52.29 \\
\hline
AN1		& PCA		& 87.18	  & 226.82 & 131.45 & 54.70			 & 738.91 & 642.74 & 94.87			& 112.50 & 17.58 \\
		& LG PCA	& 97.44	  & 92.41  & 3.86   & 80.34			 & 142.82 & 48.72  & 98.29			& 87.66	 & 1.02 \\
		& MLG PCA	& 97.44	  & 93.14  & 4.31   & 90.60			 & 132.59 & 31.71  & 98.29			& 86.20	 & 0.98 \\
\hline

\end{tabular}

\end{sidewaystable}

\begin{figure}[htb]

\begin{minipage}[b]{1.0\textwidth}
\tabcaption{Average recognition results (of 3) if we use the same expressions for training and registration (known faces), but use different expressions for recognition (NE2, SM2, AN2).}
\label{tbl_rec_ar2}
\begin{tabular}{ccccccc}
\hline
Known	& Method	&\multicolumn{5}{c}{Average recognition results} \\
\cline{3-7}
faces	&			& Cum(100) & First 1 & CMCA   & ROCA	& EER 	 \\
\hline
NE1		& PCA		& 57.55 & 81.20	& 217.94	& 118.54	& 4.56 	\\
		& LG PCA	& 14.25 & 94.30	& 109.82	& 19.73		& 1.42 	\\
		& MLG PCA	& 10.54 & 96.30	& 96.55		& 8.80		& 0.85 	\\
\hline
SM1		& PCA		& 40.46 & 68.38	& 343.22	& 254.14	& 6.84 	\\
		& LG PCA	& 20.80 & 84.62	& 160.71	& 70.86		& 2.85 	\\
		& MLG PCA	& 14.53 & 91.74	& 117.73	& 28.16		& 2.28 	\\
\hline
AN1		& PCA		& 62.68 & 78.92	& 359.41	& 263.92	& 6.55 	\\
		& LG PCA	& 10.54 & 92.02	& 107.63	& 17.87		& 1.14 	\\
		& MLG PCA	& 15.38 & 95.44	& 103.98	& 12.33		& 0.85 	\\
\hline
\end{tabular}
\end{minipage}

\begin{minipage}[b]{1.0\textwidth}
\centering \includegraphics[width=120mm]{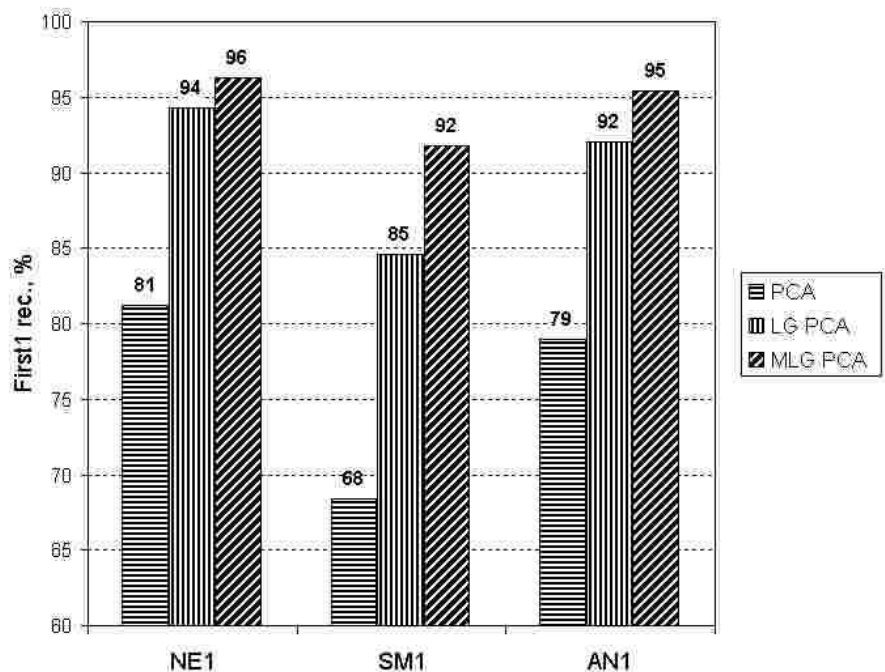}
\caption{Average first one recognition accuracy using AR database and different expressions for training.}
\label{fig_ar_first1}
\end{minipage}

\end{figure}

Average recognition results showed (Table \ref{tbl_rec_ar2}, Fig. \ref{fig_ar_first1}) that if we wish to use the same
expressions for training and registration,
and then recognize faces with different expressions, the best choice (with respect to first one recognition accuracy)
for training and registration are images with neutral (NE1) or angry (AN1) expressions, and that we should not use
faces with smiling (SM1) expressions. Also we can see that MLG PCA in most cases achieves higher average recognition
accuracy than PCA and LG PCA.

\clearpage

\subsection{Recognition experiments using FERET database}

Because the AR database is relatively small (small number of different persons), we
also performed recognition experiments using much larger FERET database \cite{Phillips1998}
containing greyscale photographs of 1196 persons.
This database was collected in 1993-1996 by the researchers from
George Mason University during the FERET (FacE REcognition Technology) program.
As far as we know, this is the largest database of face photographs (of different persons) in the world that is publicly
available for face recognition research purposes.
For training we used 1196 greyscale images (the size of each image is 256x384 pixels) from the $fa$ set of this database.
The same 1196 $fa$ images were used as a gallery (known persons), and 1195 images from the $fb$ set were used as a probe (unknown persons that we wish to recognize).
Facial images from these sets have different expressions, but these expressions are not strictly controlled.
For log-Gabor feature masking we used binary masks that were generated from the AR database's faces.

\begin{figure}[htb]
\centering \includegraphics[width=120mm]{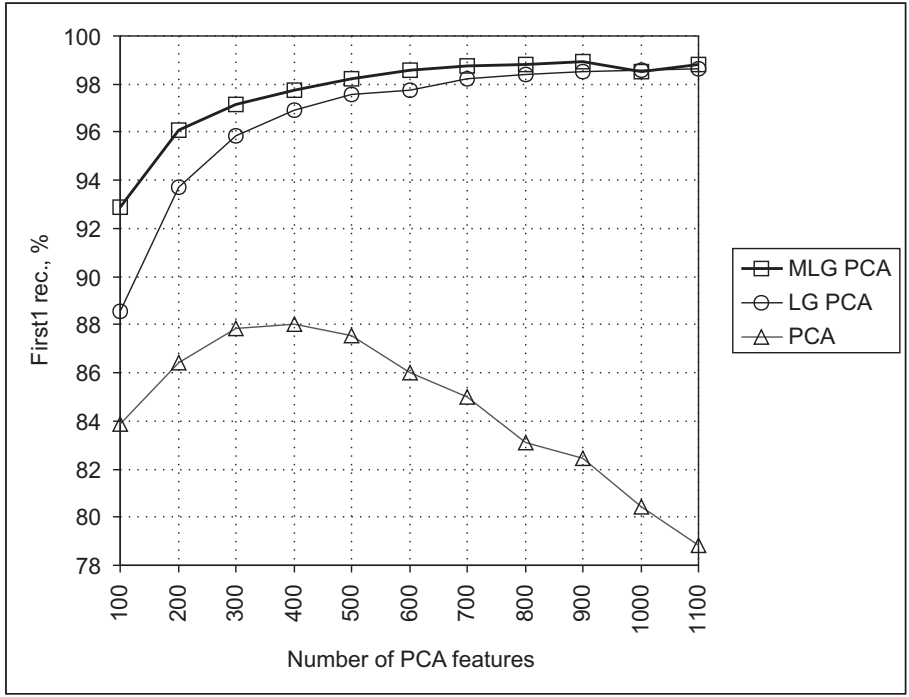}
\caption{First one recognition accuracy using FERET database and different number of PCA features.}
\label{fig_feret_first1}
\end{figure}

At first we performed recognition experiments using different number (100-1100) of PCA features
and compared the results of PCA, LG PCA, and MLG PCA methods. These results are presented in Table \ref{tbl_rec_feret1}.
The results showed that using MLG PCA or LG PCA we can achieve
much better recognition results than using traditional PCA with respect to all measured characteristics
(Cum(97)-Cum(100), first one recognition, CMCA, ROCA, EER). The results also showed that MLG PCA is better
(or not worse) than LG PCA with respect to Cum(97)-Cum(99), ROCA, EER. The MLG PCA also achieves higher first
one recognition accuracy than LG PCA. But LG PCA is slightly better than MLG PCA with respect to Cum(100) and
in some cases with respect to CMCA. First one recognition accuracies of compared methods using different number of PCA
features are also presented in Fig. \ref{fig_feret_first1}. This figure illustrates that MLG PCA achieves
higher accuracy than other compared methods. Also we can see that first one recognition accuracy of traditional PCA -based
recognition method (using cosine -based distance measure between "whitened" feature vectors)
decreases if we use larger number of PCA features. The same effect was observed in \cite{Perlibakas2004a}
using other database and image normalization procedure.

\begin{table*}[htbp]
\caption{Comparison of PCA, LG PCA and MLG PCA methods using different number of PCA features.
For experiments was used the FERET database (1196 $fa$, 1195 $fb$) and 1196 $fa$ training images.}
\label{tbl_rec_feret1}

\begin{tabular}{cccccccccc}
\hline
Method	& Feat.	& Cum(97) & Cum(98) & Cum(99) & Cum(100) & First 1 & CMCA   & ROCA	& EER 	 \\
		& num.  &         &         &         &          &         &        &       &        \\
\hline
PCA		& 100	& 1.59	& 2.93	& 6.44	& 36.87	& 83.85	& 37.27	& 28.77	& 2.26 \\
LG PCA	& 100	& 0.75	& 1.17	& 2.26	& 10.70	& 88.54	& 15.89	& 7.42	& 1.34 \\
MLG PCA	& 100	& 0.42	& 0.50	& 1.25	& 10.87	& 92.89	& 13.64	& 4.86	& 0.92 \\
\hline
PCA		& 200	& 1.17	& 2.34	& 7.19	& 84.20	& 86.44	& 40.21	& 31.59	& 2.01 \\
LG PCA	& 200	& 0.25	& 0.33	& 0.75	& 4.77	& 93.72	& 10.87	& 2.10	& 0.59 \\
MLG PCA	& 200	& 0.17	& 0.25	& 0.59	& 8.11	& 96.07	& 10.34	& 1.68	& 0.50 \\
\hline
PCA		& 300	& 1.34	& 2.84	& 5.52	& 93.65	& 87.87	& 39.36	& 29.91	& 1.92 \\
LG PCA	& 300	& 0.17	& 0.25	& 0.42	& 6.52	& 95.82	& 10.00	& 1.28	& 0.50 \\
MLG PCA	& 300	& 0.08	& 0.17	& 0.33	& 1.67	& 97.15	& 9.09	& 0.49	& 0.33 \\
\hline
PCA		& 400	& 1.67	& 3.09	& 12.12	& 92.73	& 88.03	& 52.94	& 42.79	& 2.09 \\
LG PCA	& 400	& 0.17	& 0.17	& 0.33	& 1.76	& 96.90	& 9.08	& 0.57	& 0.33 \\
MLG PCA	& 400	& 0.08	& 0.17	& 0.25	& 1.59	& 97.74	& 8.90	& 0.34	& 0.33 \\
\hline
PCA		& 500	& 2.01	& 4.10	& 16.64	& 74.50	& 87.53	& 58.73	& 48.26	& 2.34 \\
LG PCA	& 500	& 0.08	& 0.17	& 0.25	& 1.17	& 97.57	& 8.85	& 0.36	& 0.25 \\
MLG PCA	& 500	& 0.08	& 0.08	& 0.17	& 1.67	& 98.24	& 8.80	& 0.26	& 0.33 \\
\hline
PCA		& 600	& 2.59	& 5.94	& 13.55	& 82.86	& 86.03	& 67.05	& 56.21	& 2.59 \\
LG PCA	& 600	& 0.08	& 0.17	& 0.17	& 0.84	& 97.74	& 8.72	& 0.24	& 0.25 \\
MLG PCA	& 600	& 0.08	& 0.08	& 0.17	& 1.92	& 98.58	& 8.69	& 0.16	& 0.17 \\
\hline
PCA		& 700	& 4.35	& 8.86	& 22.41	& 95.07	& 85.02	& 88.53	& 77.02	& 3.18 \\
LG PCA	& 700	& 0.08	& 0.08	& 0.25	& 0.75	& 98.24	& 8.68	& 0.22	& 0.25 \\
MLG PCA	& 700	& 0.08	& 0.08	& 0.17	& 1.59	& 98.74	& 8.69	& 0.15	& 0.17 \\
\hline
PCA		& 800	& 5.35	& 10.28	& 28.51	& 93.14	& 83.10	& 109.43 & 97.12 & 3.51 \\
LG PCA	& 800	& 0.08	& 0.08	& 0.17	& 0.75	& 98.41	& 8.68	 & 0.20	 & 0.25 \\
MLG PCA	& 800	& 0.08	& 0.08	& 0.17	& 1.59	& 98.83	& 8.68	 & 0.14	 & 0.17 \\
\hline
PCA		& 900	& 7.02	& 12.63	& 38.13	& 92.89	& 82.43	& 123.39 & 110.99 & 3.93 \\
LG PCA	& 900	& 0.08	& 0.08	& 0.17	& 0.50	& 98.49	& 8.58	 & 0.14	  & 0.17 \\
MLG PCA	& 900	& 0.08	& 0.08	& 0.17	& 1.34	& 98.91	& 8.64	 & 0.13	  & 0.17 \\
\hline
PCA		& 1000	& 9.28	& 16.22	& 45.48	& 90.80	& 80.42	& 143.82 & 130.79 & 4.27 \\
LG PCA	& 1000	& 0.08	& 0.08	& 0.17	& 0.75	& 98.58	& 8.62	 & 0.14	  & 0.17 \\
MLG PCA	& 1000	& 0.08	& 0.08	& 0.17	& 0.75	& 98.49	& 8.61	 & 0.10	  & 0.17 \\
\hline
PCA		& 1100	& 12.63	& 26.25	& 48.75	& 91.47	& 78.83	& 166.00 & 152.41 & 4.77 \\
LG PCA	& 1100	& 0.08	& 0.08	& 0.17	& 0.92	& 98.66	& 8.65	 & 0.13	  & 0.17 \\
MLG PCA	& 1100	& 0.08	& 0.08	& 0.17	& 1.00	& 98.83	& 8.67	 & 0.12	  & 0.17 \\
\hline
\end{tabular}
\end{table*}

\begin{table*}[htb]
\caption{Average recognition results of 50 experiments using FERET database (1196 $fa$, 1195 $fb$), 400 $fa$ training images, 380 PCA features.}
\label{tbl_rec_feret2}

\begin{tabular}{cccccccccc}
\hline
Method	& Cum(97) & Cum(98) & Cum(99) & Cum(100) & First 1 & CMCA   & ROCA	& EER 	 \\
\hline
\hline
PCA		& 2.59	& 4.62	& 11.21	& 70.66	& 83.83	& 52.09	& 43.92	& 2.69 \\
		& $\pm$ 0.46	& $\pm$ 0.97	& $\pm$ 2.67	& $\pm$ 17.33	& $\pm$ 0.90	& $\pm$ 5.70	& $\pm$ 5.86	& $\pm$ 0.20 \\
\hline
LG PCA	& 0.23	& 0.35	& 0.73	& 5.72	& 94.45	& 10.88	& 2.31	& 0.70 \\
		& $\pm$ 0.05	& $\pm$ 0.08	& $\pm$ 0.19	& $\pm$ 2.47	& $\pm$ 0.56	& $\pm$ 0.56	& $\pm$ 0.45	& $\pm$ 0.08 \\
\hline
MLG PCA	& 0.16	& 0.20	& 0.45	& 5.60	& 96.59	& 9.97	& 1.46	& 0.56 \\
		& $\pm$ 0.03	& $\pm$ 0.04	& $\pm$ 0.11	& $\pm$ 3.55	& $\pm$ 0.40	& $\pm$ 0.53	& $\pm$ 0.43	& $\pm$ 0.07 \\
\hline

\end{tabular}
\end{table*}

Because in real life situations it is not always possible to use large number (more than 1000) of
training images, we tested our face recognition methods using 400 $fa$ training images and 380 PCA features.
We performed 50 experiments with different training sets of 400 $fa$ images
and calculated average recognition results. These training sets for all compared methods were the same.
These average results and standard deviations are presented in Table \ref{tbl_rec_feret2}.
\vbox{
The best results (Table \ref{tbl_rec_feret2}) were achieved using MLG PCA.
Even using small number of training images this method 
achieved enough high first one recognition accuracy ($>$96\%) and low EER (0.6\%).
Traditional PCA achieved 12\% lower first one accuracy and 2\% higher EER.
}

\subsection{Comparison of our results with the results of other researchers}

For comparison we used the FERET database, which has become a de facto
standard for evaluating face recognition technologies \cite{Phillips2000}.
We compared the results of different face recognition methods that were tested
using all images (not subsets) from the FERET $fa$ (1196 images) and $fb$ (1195 images) sets.
For comparison we present only the best published results of other researchers that we were able to find.
More results of other researchers could be found in \cite{Phillips2000}, \cite{NIST2001}, \cite{CSU2003}.
The selected results are summarized in Table ~\ref{tbl_rec_feret}.

\begin{sidewaystable}
\caption{Comparison of different face recognition methods using FERET database (1196 $fa$, 1195 $fb$).}
\label{tbl_rec_feret}

\begin{tabular}{lccccccccc}
\hline
Method and its authors&  Cum(97) & Cum(98) & Cum(99) & Cum(100) & First 1 & CMCA   & ROCA	& EER 	 \\
\hline
\hline
MIT 1996 (Dual PCA + Bayes MAP) \cite{Moghaddam1996}	&	0.33&	1.09&	23.33&	99.83	&	94.81 &	84.14		&	203.25 &	4.77		\\
\hline
UMD 1997 (PCA+LDA) \cite{Etemad1997}, \cite{Zhao1998}	&	0.17&	0.33&	0.84&	75.92	&	96.23 &	18.91		&	14.37 &	1.09		\\
\hline
USC 1997 (EBGM) \cite{Wiskott1997}, \cite{Okada1998}&	0.25&	0.33&	3.09&	50.67	&	94.98 &	27.94		&	57.52 &	2.51		\\
\hline
CSU EBGM Standard \cite{Bolme2003} &	1.34&	2.42&	9.2	&	37.54	&	88.37 &	34.26	&	-		&	-		\\
\hline
CSU EBGM Optimised \cite{Bolme2003} &	-	&	-	&	-	&	-	&	89.80 &	-		&	-		&	-		\\
\hline
CSU PCA MahCosine \cite{CSU2003} &	2.26&	4.43&	10.28&	60.45	&	85.27 &	48.90		&	-		&	-		\\
\hline
Gabor features \cite{Kepenekci2002} &	-	&	-	&	-	&	-		&	96.30 &	-			&	-		&	-		\\
\hline
Haar+AdaBoost \cite{Jones2003} &	-	&	$\sim$0.42	&	$\sim$1.17	&	-		&	$\sim$94.00 &	-			&	- &	$\sim$1.00				\\
\hline
Gabor+AdaBoost \cite{Yang2004} &	-	&	-	&	-	&	-		&	$\sim$95.20 &	-			&	-		&	-		\\
\hline

Our MLG PCA (train 400 fa, 380 PCA feat.)	& 0.16 & 0.20 & 0.45 & 5.60 &	96.59 &	9.97 	&	1.64 &	0.56			\\
average values of 50 experiments            &      &      &       &       &       &         &	      &				\\
\hline
Our MLG PCA (train 1196 fa, 300 PCA feat.)	& 0.08 & 0.17 & 0.33 & 1.67 &	97.15 &	9.09 	&	0.49 &	0.33			\\
\hline
Our MLG PCA (train 1196 fa, 900 PCA feat.)	& 0.08 & 0.08 & 0.17 & 1.34 &	98.91 &	8.64 	&	0.13 &	0.17			\\
\hline

\end{tabular}
\end{sidewaystable}

Now we will briefly describe the methods that we selected for comparison.
MIT 1996 (Massachusetts Institute of Technology) method \cite{Moghaddam1996} 
uses dual (intrapersonal and extrapersonal) PCA and Bayesian MAP (maximum a posteriori) similarity measure.
For training are used image pairs of the same and of different persons.
UMD 1997 (University of Maryland) method \cite{Etemad1997}, \cite{Zhao1998}
uses a combination of Principal Component Analysis (PCA) and Linear Discriminant Analysis (LDA) methods and
weighted Euclidean distance between feature vectors. For LDA training are used multiple ($>2$) images of the same person.
For training are used more than 1000 images of more than 400 persons. For feature extraction are used 300 eigenvectors.
USC 1997 (University of Southern California)
Elastic Bunch Graph Matching (EBGM) method \cite{Wiskott1997}, \cite{Okada1998}
detects specified facial features (48 graph nodes) and extracts Gabor Jets using 40 Gabor filters (5 scales, 8 orientations).
Geometrical relationships between features and the values of Gabor Jets (at graph nodes) are used for comparison of faces.
At first this method is trained using 70 facial images with manually selected features, and then
these features are detected automatically. For comparison are used more than 1900 features.
CSU EBGM Standard \cite{CSU2003} and CSU EBGM Optimised \cite{Bolme2003}
face recognition methods were developed at Colorado State University.
These methods use the same theoretical background as the method of USC, but different features and
different training images. For training are used 70 facial images with manually selected feature landmarks.
Features are extracted using 80 Gabor filters (8 orientations, 5 frequencies, 2 phases).
For recognition are used graphs with 80 nodes (25 landmarks, 55 interpolated points).
The total number of features is larger than 6000. It is important to note that EBGM -based methods
for comparison usually use a whole human head (not only the internal part of the face without hair and facial contour).
CSU PCA MahCosine method \cite{CSU2003} uses traditional PCA and cosine -based distance measure between "whitened"
feature vectors, 501 training images, and 300 PCA features.
Gabor features -based method \cite{Kepenekci2002} extracts features by using
40 Gabor filters and then uses sliding window -based algorithm for finding high-energized points.
Then multiple complex-valued feature vectors are constructed by storing the coordinates of the points and
40 filter responses at these points (each feature vector has 42 components). Recognition is performed by calculating
similarity measures between multiple vectors and then by calculating the overall similarity measure.
Haar+AdaBoost method \cite{Jones2003} uses Haar-like features and adaptive boosting -based training and feature selection.
This method is trained using 398 pairs of images from the FERET $fa$ and $fb$ sets. After training are selected 400 features.
Gabor+AdaBoost method \cite{Yang2004} for feature extraction uses Gabor filters (5 scales, 8 orientations).
Adaptive boosting -based training is used in order to reduce the number of intrapersonal and extrapersonal features.
For recognition are selected 700 features.

The comparison showed (Table ~\ref{tbl_rec_feret}) that
our MLG PCA method can achieve higher first one recognition accuracy (up to 98.91\% using 1196 $fa$ training images and 900 PCA features)
and lower EER (0.17\%) than many other compared methods.
Even using only 400 $fa$ training images and 380 PCA features our MLG PCA method achieves enough
high recognition accuracy (First 1 = 96.59\%, EER = 0.56\%).
The best methods of other researchers are
Gabor features \cite{Kepenekci2002} (First 1 = 96.30\%),
UMD 1997 (PCA+LDA) \cite{Etemad1997}, \cite{Zhao1998}	(First 1 = 96.23\%, EER = 1.09\%),
Gabor+AdaBoost \cite{Yang2004} (First 1 = 95.20\%),
and Haar+AdaBoost \cite{Jones2003} (First 1 = 94\%, EER = 1\%).
CMCA values of our method are about two times smaller than of other methods,
and ROCA values of our method are also several times smaller than of other methods.
In order to achieve 99\% cumulative recognition rate (using 900 features)
we need to extract from the database only 0.17\% of images (that is 1196*0.17/100 = 2 images), 
and in order to achieve 100\% cumulative recognition rate we need to extract 16 images (1.34\%).
Using CSU EBGM Standard method we need to extract 449 images of 1196 (37.54\%) in order to achieve 100\% cumulative
recognition accuracy, and using
USC 1997 (EBGM) \cite{Wiskott1997}, \cite{Okada1998} method we need to extract 606 images (50.67\%).
Using the proposed method we also can achieve $>$90\% recognition accuracy using larger windows
than 4x4 pixels and smaller number of log-Gabor features ($<$10000), but we prefer to use larger number of log-Gabor features (15120)
and then reduce the number of features to several hundreds by using PCA,
because it allows to achieve $>$96\% first one recognition accuracy.
Because our traditional PCA using small number of training images (400 $fa$) and features (380 PCA features)
achieves similar recognition accuracy (84\%) as CSU's \cite{CSU2003} implementation of the same method
(CSU PCA MahCosine \cite{CSU2003}, 501 mostly $fa$ training images, 300 PCA features, 85\% first one accuracy),
we decided to test if it is possible to achieve high recognition accuracy using
only log-Gabor features and different distance measures (Euclidean, Manhattan, cosine-based) without using PCA.
But the results of these experiments showed that MLG PCA achieves much higher recognition accuracy.

\clearpage

\section{Conclusions and future work}

In this article we proposed a novel face recognition method based on masked log-Gabor features and Principal Component Analysis.
The experiments with the AR and FERET databases showed that using the proposed MLG PCA method we can increase the accuracy
of recognizing faces with different expressions.
The experiments showed that using log-Gabor features, expression masking, sliding window -based feature selection method,
Principal Component Analysis, "whitening", and cosine -based distance measure we can achieve very high recognition accuracy
(98.91\%) and low Equal Error Rate (0.17\%) with the FERET database containing facial photographs of 1196 persons.
The results of our algorithm are among the best results that were ever achieved using this database.

In the future we are going to investigate other methods of extracting and comparing facial regions in order
to achieve higher recognition accuracy (of faces with different expressions) than we already achieved.
Also we are going to investigate the possibilities to optimise the parameters of log-Gabor filters in order
to increase the accuracy of the proposed method and reduce the number of features.

\section{Acknowledgements}

Portions of the research in this paper use the FERET database of facial images collected under the FERET program.


\end{document}